\documentclass[10pt,twocolumn,letterpaper]{article}

\usepackage{iccv}
\usepackage{times}
\usepackage{epsfig}
\usepackage{graphicx}
\usepackage{amsmath}
\usepackage{amssymb}
\usepackage{algorithm} 
\usepackage{algorithmic} 
\usepackage{multirow}
\usepackage{array}
\usepackage{tabularx}
\usepackage{url}

\usepackage[pagebackref=true,breaklinks=true,letterpaper=true,colorlinks,bookmarks=false]{hyperref}

\iccvfinalcopy 

\def\iccvPaperID{2120} 

\ificcvfinal\fi
\begin{document}

\title{Dynamic Label Graph Matching for Unsupervised Video Re-Identification}

\author{Mang Ye$^{1}$, Andy J Ma$^{1}$, Liang Zheng$^{2}$, Jiawei Li$^{1}$, Pong C Yuen$^{1}$ \\
$^{1}$ Hong Kong Baptist University \quad \quad $^{2}$ University of Technology Sydney \\
{\tt\small \{mangye,andyjhma,jwli,pcyuen\}@comp.hkbu.edu.hk, liangzheng06@gmail.com}
}

\maketitle
\thispagestyle{empty}

\begin{abstract}

Label estimation is an important component in an unsupervised person re-identification (re-ID) system. This paper focuses on cross-camera label estimation, which can be subsequently used in feature learning to learn robust re-ID models. Specifically, we propose to construct a graph for samples in each camera, and then graph matching scheme is introduced for cross-camera labeling association. While labels directly output from existing graph matching methods may be noisy and inaccurate due to significant cross-camera variations, this paper propose a dynamic graph matching (DGM) method. DGM iteratively updates the image graph and the label estimation process by learning a better feature space with intermediate estimated labels. DGM is advantageous in two aspects: 1) the accuracy of estimated labels is improved significantly with the iterations; 2) DGM is robust to noisy initial training data. Extensive experiments conducted on three benchmarks including the large-scale MARS dataset show that DGM yields competitive performance to fully supervised baselines, and outperforms competing unsupervised learning methods.\footnote{Code is available at \url{www.comp.hkbu.edu.hk/\%7e mangye/}}

\end{abstract}

\section{Introduction}
Person re-identification (re-ID), a retrieval problem in its essence \cite{zheng2017sift,tmm16rank,arxiv17survey}, aims to search for the queried person from a gallery of disjoint cameras. In recent years, impressive progress has been reported in video based re-ID \cite{cvpr16top,cvpr16video,eccv16mars}, because video sequences provide rich visual and temporal information and can be trivially obtained by tracking algorithms \cite{tip15track,aaai17track} in practical video surveillance applications. Nevertheless, the annotation difficulty limits the scalability of supervised methods in large-scale camera networks, which motivates us to investigate an unsupervised solution for video re-ID.


\begin{figure}[t]
  \centering
  \includegraphics[width=1\linewidth]{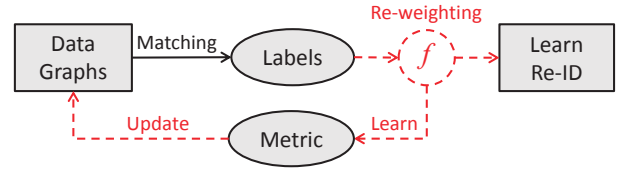}\\
  \caption{\small{Pipeline Illustration. Graph matching is conducted after constructing a graph for samples in each camera to obtain the intermediate labels. Instead of using the labels directly, label re-weighting is introduced to handle the noisy intermediate labels. Iteratively, the graph is updated, labels are estimated, and distance metrics are learnt.}}\label{fig:idea}
\end{figure}

The difference between unsupervised learning and supervised learning consists in the availability of labels. Considering the good performance of supervised methods, an intuitive idea for unsupervised learning is to estimate re-ID labels as accurately as possible. In previous works, part from directly using hand-crafted descriptors \cite{eccv14yang,cvpr15lomo,cvpr16gog,iccv15des}, some other unsupervised re-ID methods focus on finding shared invariant information (saliency \cite{cvpr13saliency} or dictionary \cite{eccv16un,cvpr16un}) among cameras. Deviating from the idea of estimating labels, these methods \cite{cvpr13saliency,eccv16un,cvpr16un} might be less competitive compared with the supervised counterparts. Meanwhile, these methods also suffer from large cross-camera variations. For example, salient features are not stable due to occlusions or viewpoint variations. Different from the existing unsupervised person re-ID methods, this paper is based on a more customized solution, \ie, cross-camera label estimation. In other words, we aim to mine the labels (matched or unmatched video pairs) across cameras. With the estimated labels, the remaining steps are exactly the same with supervised learning.

To mine labels across cameras, we leverage the graph matching technique (\eg, \cite{pami16gm}) by constructing a graph for samples in each camera for label estimation. Instead of estimating labels independently, the graph matching approach has shown good property in finding correspondences by minimize the globally matching cost with intra-graph relationship.
Meanwhile, label estimation problem for re-ID task is to link the same person across different cameras, which perfectly matches the graph matching problem by treating each person as a graph node.
However, labels directly estimated by existing graph matching are very likely to be inaccurate and noisy due to the significant appearance changes across cameras. So a fixed graph constructed in the original feature space usually does not produce satisfying results. Moreover, the assumption that the assignment cost or affinity matrix is fixed in most graph matching methods may be unsuitable for re-ID due to large cross-camera variations \cite{ijcv12gm,iccv15joint,cho10rrwm,pami16gm}.

In light of the above discussions, this paper proposes a dynamic graph matching (DGM) method to improve the label estimation performance for unsupervised video re-ID (the main idea is shown in Fig. \ref{fig:idea}). Specifically, our pipeline is an iterative process. In each iteration, a bipartite graph is established, labels are then estimated, and then a discriminative metric is learnt. Throughout this procedure, labels gradually become more accurate, and the learnt metric more discriminative. Additionally, our method includes a label re-weighting strategy which provides soft labels instead of hard labels, a beneficial step against the noisy intermediate label estimation output from graph matching.

The main contributions are summarized as follows:
\begin{itemize}
\item We propose a dynamic graph matching (DGM) method to estimate cross-camera labels for unsupervised re-ID, which is robust to distractors and noisy initial training data. The estimated labels can be used for further discriminative re-ID models learning.
\item Our experiment confirms that DGM is only slightly inferior to its supervised baselines and yields competitive re-ID accuracy compared with existing unsupervised re-ID methods on three video benchmarks.
 \end{itemize}
\section{Related Work}
\textbf{Unsupervised Re-ID.}
Since unsupervised methods could alleviate the reliance on large-scale supervised data, a number of unsupervised methods have been developed. Some transfer learning based methods \cite{cvpr16un,iccv13ma,cvpr17un} are proposed. Andy \emph{et al.} \cite{iccv13ma} present a multi-task learning method by aligning the positive mean on the target dataset to learn the re-ID models for the target dataset. Peng \emph{et al.} \cite{cvpr16un} try to adopt the pre-trained models on the source datasets to estimate the labels on the target datasets. Besides that, Zhao \emph{et al.} \cite{cvpr13saliency} present a patch based matching method with inconsistent salience for re-ID. An unsupervised cross dataset transfer learning method with graph Laplacian regularization terms is introduced in \cite{cvpr16un}, and a similar constraint with graph Laplacian regularization term for dictionary learning is proposed in \cite{eccv16un} to address the unsupervised re-ID problem. Khan \emph{et al.} \cite{avss16reid} select multiple frames in a video sequence as positive samples for unsupervised metric learning, which has limited extendability to the cross-camera settings.

Two main differences between the proposed method and previous unsupervised re-ID methods are summarized. Firstly, this paper estimates labels with graph matching to address the cross-camera variation problem instead of directly learning an invariant representation.  Secondly, output estimated labels of dynamic graph matching can be easily expanded with other advanced supervised learning methods, which provides much flexibility for practical applications in large-scale camera network.

Two contemporary methods exists \cite{iccv17labeling,fan2017unsupervised} which also employ the idea of label estimation for unsupervised re-ID. Liu \emph{et al.} \cite{iccv17labeling} use a retrieval method for labeling, while Fan \emph{et al.} \cite{fan2017unsupervised} employ $k$-means for label clustering.


%

\textbf{Graph Matching for Re-ID.}
Graph matching has been widely studied in many computer vision tasks, such as object recognition and shape matching \cite{pami16gm}. It has shown superiority in finding consistent correspondences in two sets of features in an unsupervised manner. The relationships between nodes and edges are usually represented by assignment cost matrix \cite{ijcv12gm,iccv15joint} or affinity matrix \cite{cho10rrwm,pami16gm}. Currently graph matching mainly focuses on optimizing the matching procedure with two fixed graphs. That is to say, the affinity matrix is fixed first, and then graph matching is formulated as linear integer programs \cite{iccv15joint} or quadratic integer programs \cite{ijcv12gm}. Different from the literature, the graph constructed based on the original feature space is sub-optimal for re-ID task, since we need to model the camera variations besides the intra-graph deformations. Therefore, we design a dynamic graph strategy to optimize matching. Specifically, partial reliable matched results are utilized to learn discriminative metrics for accurate graph matching in each iteration.


Graph matching has been introduced in previous re-ID works which fall into two main categories. (1) Constructing a graph for each person by representing each node with body parts \cite{iccv13gm} or local regions \cite{tcsvt16gm}, and then a graph matching procedure is conducted to do re-identification. (2) Establishing a graph for each camera view, Hamid et al. \cite{cvpr16graph} introduces a joint graph matching to refine final matching results. They assume that all the query and gallery persons are available for testing, and then the matching results can be optimized by considering their joint distribution. However, it is hard to list a practical application for this method, since only the query person is available during testing stage in most scenarios. Motivated by \cite{cvpr16graph}, we construct a graph for each camera by considering each person as a node during the training procedure. Subsequently, we could mine the positive video pairs in two cameras with graph matching.


\section{Graph Matching for Video Re-ID}\label{sec:gm}
Suppose that unlabelled graph $\mathcal{G_A}$ contains $m$ persons, which is represented by $[\mathcal{A}] = \{\mathbf{x}^{i}_{a}|i = 1,2,\cdots,m\}$ for camera A, and another graph $\mathcal{G_B}$ consists of $n$ persons denoted by $[\mathcal{B}]_0 = \{\mathbf{x}^{j}_{b}|j = 0,1,2,\cdots,n\}$ for camera B. Note that $[\mathcal{B}]_0$ contains another 0 element besides the $n$ persons. The main purpose is to model the situation that more than one person in $\mathcal{G_A}$ cannot find its correspondences in $\mathcal{G_B}$, \ie allowing person-to-dummy assignments. To mine the label information across cameras, we follow \cite{iccv15joint} to formulate it as a binary linear programming with linear constraints:
\begin{equation}\label{eq:match}
\begin{split}
&G(\mathbf{y}) = \arg \mathop {\min }\limits_{Y} C^{T}\mathbf{y} \\
s.t.\quad &\forall i \in [\mathcal{A}],\forall j \in [\mathcal{B}]_{0}:y^{j}_{i}\in\{0,1\},\\
&\forall j \in [\mathcal{B}]_{0}:{\sum\limits_{i \in [\mathcal{A}]} {y_i^j \le 1} },\\
&\forall i \in [\mathcal{A}]:{\sum\limits_{j \in [\mathcal{B}]_{0}} {y_i^j = 1} },
\end{split}
\end{equation}
where $\mathbf{y}= \{y_i^j\}\in \mathbb{R}^ {m(n+1)\times 1}$ is an assignment indicator of node $i$ and $j$, representing whether $i$ and $j$ are the same person ($y_i^j =1$) or not ($y_i^j =0$). $C = \{C(i,j)\}$ is the assignment cost matrix with each element illustrating the distance of node $i$ to node $j$. The assignment cost is usually defined by node distance like $C(i,j) = Dist(\mathbf{x}^{i}_{a},\mathbf{x}^{j}_{b})$, as done in \cite{cvpr16graph}. Additionally, some geometry information is added in many feature point matching models \cite{ijcv12gm}.

For video re-ID, each node (person) is represented by a set of frames. Therefore, \textit{Sequence Cost} (${C_S}$) and \textit{Neighborhood Cost} ($C_{N}$) are designed as the assignment cost in the graph matching model for video re-ID under a certain metric. The former cost penalizes matchings with mean set-to-set distance, while the latter one constrains the graph matching with within-graph data structure. The assignment cost between person $i$ and $j$ is then formulated as a combination of two costs with a weighting parameter $\lambda$ in a log-logistic form:
\begin{equation}\label{eq:assigncost}
C = log(1+ e^ {(C_{S} + \lambda C_{N})}).
\end{equation}

\textbf{Sequence Cost}. The sequence cost $C_{S}$ penalizes the matched sequences with the sequence difference. Under a discriminative metric $M$ learnt from frame-level features, the average set distance between video sequences $\{x_a^i\}$ and $\{x_b^j\}$ is defined as the sequence cost, \ie,
\begin{equation}\label{eq:Seqcost}
C_{S}(i,j) = \frac{1}{|\{x_a^i\}||\{x_b^j\}|}\sum_{} \sum_{}  D_M( x_a^{i_m},x_b^{j_n}).
\end{equation}
\begin{figure}[t]
  \centering
  \includegraphics[width=6.7cm,height = 4cm]{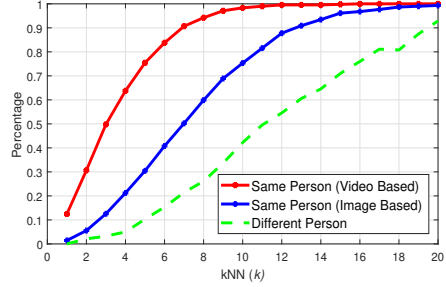}\\
  \caption{\small{Illustration of the neighborhood similarity. With various values of $k$, we record the percentages of having intersection of same (different) person's kNN under two different cameras. The \textit{Same Person (Video-based)} refers to video re-ID task in which one person have multiple person images. \textit{Same Person (Image-based)} denotes the image based re-ID task in which each person only have single image per camera.}}\label{fig:illustration}
\end{figure}

\textbf{Neighborhood Cost}. The neighborhood cost $C_{N}$ models the within camera data structure with neighborhood similarity constraints. Specifically, the correctly matched person pair's neighborhood under two cameras should be similar \cite{mmm15,mm15rank}. A primarily experiment on PRID2011 dataset with features in \cite{iccv15des} is conducted to justify this point. Results shown in Fig. \ref{fig:illustration} illustrates that the percentages of the same person having common neighbors are much larger than that of different persons. It means that the same person under two different cameras should share similar neighborhood \cite{tmm16cross}. Moreover, compared with image-based re-ID, the neighborhood similarity constraints for video-based re-ID are much more effective. It verifies our idea to integrate the neighborhood constraints for graph matching in video re-ID, which could help to address the camera camera variations. The neighborhood cost $C_N$ penalizes the neighborhood difference between all matched sequences, which is formulated by,
\begin{equation}\label{eq:Nieghcost}
\begin{split}
& C_{N}(i,j) =  \frac{1}{|\mathcal{N}^i_{a}||\mathcal{N}^j_{b}|} \sum_{\bar{x}_a^{i'} \in \mathcal{N}^i_{a}} \sum_{{\bar{x}_b^{j'}} \in \mathcal{N}^j_{b}}{D_M( \bar{x}_a^{i'}, \bar{x}_b^{j'})}\\
& s.t. ~~\mathcal{N}^i_{a}(i,k) = \left\{ \bar{x}_a^{i'} | D_M(\bar{x}_a^i, \bar{x}_a^{i'}) < k \right\}, \\
& ~~~~~~~\mathcal{N}^j_{b}(j,k) = \left\{ \bar{x}_b^{j'} | D_M(\bar{x}_b^j, \bar{x}_b^{j'}) < k \right\},
\end{split}
\end{equation}
where $\mathcal{N}^i_{a}$ and $\mathcal{N}^j_{b}$ denote the neighborhood of person $i$ in camera $A$ and person $j$ in camera $B$, $k$ is the neighborhood parameter. For simplicity, a general kNN method is adopted in our paper, and $k$ is set as 5 for all experiments. Meanwhile, a theoretical analysis of the neighborhood constraints is presented. Let $\bar{x}_a^p$ be a neighbor of person $i$ in camera A and $\bar{x}_b^q$ be its neighbor in camera B. From the geometry perspective, we have
\begin{equation}\label{eq:setNpres}
\begin{split}
D_M(\bar{x}_a^p,\bar{x}_b^q)\leq D_M(\bar{x}_a^p,\bar{x}_a^i) + D_M(\bar{x}_b^i,\bar{x}_b^q) +D_M(\bar{x}_a^i,\bar{x}_b^i).
\end{split}
\end{equation}
Since $\bar{x}_a^p$ and $\bar{x}_b^q$ are the neighbors of $\bar{x}_a^i$ and $\bar{x}_b^i$, respectively, $D_M(\bar{x}_a^p,\bar{x}_a^i)$ and $D_M(\bar{x}_b^i,\bar{x}_b^q) $ are small positive numbers. On the other hand, $D_M(\bar{x}_a^i,\bar{x}_b^i)$ is also a small positive under a discriminative metric $D_M$. Thus, the distance between two neighbors $\bar{x}_a^p$ and $\bar{x}_b^q$ is small enough, \ie,
\begin{equation}\label{eq:assum}
\begin{split}
D_M(\bar{x}_a^p,\bar{x}_b^q)\leq\varepsilon.
\end{split}
\end{equation}
\section{Dynamic Graph Matching}\label{sec:metric}
\begin{figure}[t]
  \centering
  \includegraphics[width=0.98\linewidth]{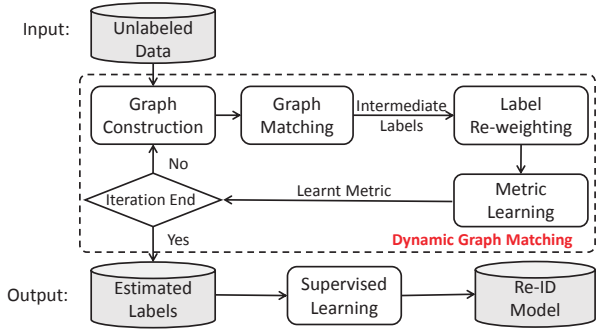}\\
  \caption{\small{Block diagram of the proposed approach. The estimated labels and learnt metric are updated in an iterative manner. }}\label{fig:frame}
\end{figure}
A number of effective graph matching optimization methods could be adopted to solve the matching problem. After that, an intuitive idea to solve unsupervised video re-ID is learning a re-identification model based on the output of graph matching. However, there still remains two obvious shortcomings:
 \begin{itemize}
\item Since existing graphs are usually constructed in the original feature space with fixed assignment cost, it is not good enough for re-ID problem due to the large cross camera variations. Therefore, we need to learn a discriminative feature space to optimize the graph matching results.
\item The estimated labels output by graph matching may bring in many false positives and negatives to the training process. Moreover, the imbalanced positive and negative video pairs would worsen this situation further. Therefore, it is reasonable to re-encode the weights of labels for overall learning, especially for the uncertain estimated positive video pairs.
 \end{itemize}
To address above two shortcomings, a dynamic graph matching method is proposed. It iteratively learns a discriminative metric with intermediate estimated labels to update the graph construction, and then the graph matching is improved. Specifically, a re-weighting scheme is introduced for the estimated positive and negative video pairs. Then, a discriminative metric learning method is introduced to update the graph matching. The block diagram of the proposed method is shown in Fig. \ref{fig:frame}.
\subsection{Label Re-weighting}
This part introduces the designed label re-weighting scheme. Note that the following re-weighting scheme is based on the output ($\mathbf{y}$) of optimization problem Eq.~\ref{eq:match}. $y_i^j \in \{0,1\}$ is a binary indicator representing whether $i$ and $j$ are the same person ($y_i^j =1$) or not ($y_i^j =0$).

\textbf{Positive Re-weighting}. All $ y_i^j=1$ estimated by graph matching are positive video pairs. Since the labels are uncertain, it means that considering all $ y_i^j=1$ equally is unreasonable. Therefore, we design a soft label $l_+(i,j)$ encoded with a Gaussian kernel for $ y_i^j=1$,
\begin{equation}\label{eq:pos}
l_+(i,j) = \left\{ {\begin{array}{ll}
{e^{-C(i,j)}  ,} & {\textrm{if $C(i,j) < \lambda_+ $}}\\
{0,}   & {\textrm{others}}
\end{array}} \right.
\end{equation}
where $\lambda_+$ is the pre-defined threshold. $C$ means the assignment cost computed in Eq. \ref{eq:assigncost} in current iteration. In this manner, the positive labels ($y =1$) are converted into soft labels, with smaller distance assigned larger weights while larger distance with smaller weights. Meanwhile, the filtering strategy could reduce the impact of false positives.

\begin{figure}[t]
  \centering
  \includegraphics[width=9cm,height = 5cm]{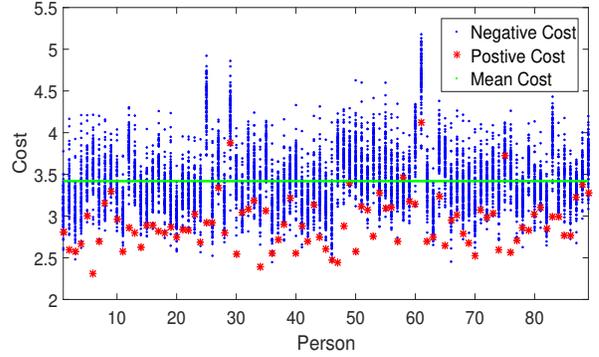}\\
  \caption{\small{Illustration about the choice of $\lambda_+$ in Eq. \ref{eq:pos} and $\lambda_-$ in Eq. \ref{eq:neg} on the PRID-2011 dataset. It is shown that most positive pair costs are smaller than the mean cost, while cost larger than mean cost is likely to be negative sample pairs. }}\label{fig:mean}
\end{figure}

\textbf{Negative Re-weighting}. Since abundant negative video pairs exist in video re-ID task compared with positive video pairs, some hard negative are selected for efficient training, $l_-(i,j)$ for all $y_i^j =0$ is defined as
\begin{equation}\label{eq:neg}
l_-(i,j) = \left\{ {\begin{array}{ll}
 {-1 ,} & {\textrm{if $C(i,j) < \lambda_- $}}\\
{0,}   & {\textrm{others,}}
\end{array}} \right.
\end{equation}
where $\lambda_-$ is the pre-defined threshold. Considering both Eq. \ref{eq:pos} and Eq. \ref{eq:neg}, we define $\lambda_+ = \lambda_- = c_m$ based on the observation shown in Fig \ref{fig:mean}. $c_m$ denotes the mean of $C$, which would be quite efficient. Thus, the label re-weighting scheme is refined by
\begin{equation}\label{eq:rew}
l(i,j) = \left\{ {\begin{array}{ll}
{e^{-C(i,j)} * y_i^j,} & {\textrm{if $0 < y_i^jC(i,j) <   c_m $}}\\
{ 0, } &{\textrm{if $C(i,j) > c_m $}}\\
{ -1, } &{\textrm{$others $}.}\\
\end{array}} \right.
\end{equation}
The label re-weighting scheme has the following advantages: (1) for positive video pairs, it could filter some false positives and then assign different positive sample pairs different weights; (2) for negative video pairs, a number of easy negatives would be filtered. The re-weighing scheme is simple but effective as shown in the experiments.

\subsection{Metric Learning with Re-weighted Labels}
With the label re-weighting scheme, we could learn a discriminative metric similar to many previous supervised metric learning works. We define the loss function by log-logistic metric learning as done in \cite{iccv15liao}, \ie,
\begin{equation}\label{eq:metric1}\small
 {f^*_M}(\bar{x}^{i}_{a},\bar{x}^{j}_{b}) = \log (1 + e^{l(i,j)(D_{M}(\bar{x}^{i}_{a},\bar{x}^{j}_{b})-c_0)}),
\end{equation}
where $c_0$ is a positive constant bias to ensure $D_M$ has a lower bound. It is usually defined by the average distance between two cameras. The function $D_{M}$ denotes the distance of $\bar{x}^{i}_{a}$ and $\bar{x}^{j}_{b}$ under the distance metric $M$, which is defined by $D_{M}(\bar{x}^{i}_{a},\bar{x}^{j}_{b}) = (\bar{x}^{i}_{a} - \bar{x}^{j}_{b})^{T}M(\bar{x}^{i}_{a} - \bar{x}^{j}_{b})$. We choose the first-order statistics $\bar{x}^{i}_{a}$ and $\bar{x}^{j}_{b}$ to represent each person as done in \cite{ijcai16video,cvpr16top}.

By summing up all of sequence pairs, we obtain the probabilistic metric learning problem under an estimated $\mathbf{y}$ formulated by,
\begin{equation}\label{eq:probmetric}
F(M;\mathbf{y}) = \sum \nolimits ^{m}_{i=1} \sum \nolimits ^{n}_{j=1} \omega _{ij} {f^*_M} (\bar{x}^{i}_{a},\bar{x}^{j}_{b}),
\end{equation}
where $\omega _{ij}$ is a weighting parameter to deal with the imbalanced positive and negative pairs. The weights $\omega _{ij}$ are caculated by $\omega _{ij} = \frac{1}{|\{l(i,j)|l(i,j)>0\}|}$ if $l(i,j)>0$, and $\omega _{ij} = \frac{1}{|\{l(i,j)|l(i,j)=-1\}|}$ if $l(i,j)=-1$, where $|\cdot|$ denotes the number of candidates in the set. Note that some uncertain pairs are assigned with label $l(i,j) = 0$ without affecting the overall metric learning. The discriminative metric can be optimized by minimizing Eq. \ref{eq:probmetric} using existing accelerated proximal gradient algorithms (\eg, \cite{apg09,iccv15liao,wang17Statistical}).

\begin{algorithm}[t]
\caption{Dynamic Graph Matching (DGM)}
\label{alg:}
\begin{algorithmic}[1]
    \renewcommand{\algorithmicrequire}{\textbf{Input:}}
    \renewcommand\algorithmicensure {\textbf{Output:}}
    \REQUIRE ~~\ Unlabelled features $\mathbf{X}_a, \mathbf{X}_b$, $M^0 = \mathbb{I}$.
    \STATE Compute $C^0$ with Eq. \ref{eq:assigncost};
    \STATE Solve Eq. \ref{eq:match} to get $\mathbf{y}^0$ and $G^0$;
    \FOR{$t = 1$ to $maxIter$}
    \STATE Label Re-weighting $l^t$ with Eq. \ref{eq:rew};
    \STATE Update $M^t$ with Eq. \ref{eq:probmetric} as done in \cite{iccv15liao};
    \STATE Update cost matrix $C^t$ with Eq. \ref{eq:assigncost};
    \STATE Solve Eq. \ref{eq:match} to get $\mathbf{y}^t$;
    \IF {$G^t \geq G^{t-1}$}
    \STATE  $\mathbf{y}^t = \mathbf{y}^{t-1}$;
    \ENDIF
    \IF {converge}
    \STATE  break;
    \ENDIF
    \ENDFOR
    \ENSURE ~~\ Estimated labels $\mathbf{y}$, learnt metric $M$ .
\end{algorithmic}
\end{algorithm}
\subsection{Iterative Updating}\label{sec:iter}
 With the label information estimated by graph matching, we could learn an improved metric by selecting high-confident labeled video pairs. By utilizing the learnt metric, the assignment cost of Eq. \ref{eq:Seqcost} and Eq. \ref{eq:Nieghcost} could be dynamically updated for better graph matching in a new iteration. After that, better graph matching could provide more reliable matching results, so as to improve the previous learnt metric. Iteratively, a stable graph matching result is finally achieved by a discriminative metric. The matched result could provide label data for further supervised learning methods. Meanwhile, a distance metric learnt in an unsupervised way could also be directly adopted for re-ID. The proposed approach is summarized in Algorithm \ref{alg:}.


\textbf{Convergence Analysis.} Note that we have two objective functions $F$ and $G$ optimizing $\mathbf{y}$ and $M$ in each iteration. To ensure the overall convergence of the proposed dynamic graph matching, we design a similar strategy as discussed in \cite{eccv12graph}. Specifically, $M$ can be easily optimized by choosing a suitable working step size $\eta\leq L$, where $L$ is the Lipschitz constant of the gradient function $\bigtriangledown F(M,\mathbf{y})$. Thus, it could ensure $F(M^{t};\mathbf{y}^{t-1})\leq F(M^{t-1};\mathbf{y}^{t-1})$, a detailed proof is shown in \cite{apg09}. For $\mathbf{y}^t$ at iteration $t$, we constrain the updating procedure by keep on updating the assignment cost matrix $C^t$ until getting a better $\mathbf{y}$ which satisfies $G(M^{t};\mathbf{y}^{t})\leq G(M^{t};\mathbf{y}^{t-1})$, similar proof can be derived from \cite{eccv12graph}. By constrain the updating procedure, it could satisfy the criteria $G^{t}(\mathbf{y};M) + F^{t}(M;\mathbf{y}) \leq G^{t-1}(\mathbf{y};M) + F^{t-1}(M;\mathbf{y})$. This is validated in our experiments as discussed in Section \ref{sec:self}. Particularly, the proposed method converges steadily.

\textbf{Complexity Analysis.} In the proposed method, most computational costs focus on the iterative procedure, since we need to conduct the graph matching with Hungarian algorithm at each iteration. We need to compute the sequence cost $O(n^2)$ and neighborhood cost $O(kn+n^2)$ for each camera, and then graph matching time complexity is $O(n^3)$. Updating $M$ with accelerated proximal gradient is extremely fast as illustrated in \cite{apg09}. However, the proposed method is conducted offline to estimate labels, which is suitable for practical applications. During the online testing procedure, we only need to compute the distance between the query person $p$ and the gallery persons with the learnt re-identification model. The distance computation complexity is $O(n)$ and ranking complexity is $O(n\log n)$, which is the same as existing methods \cite{cvpr16top,iccv15liao}.

\section{Experimental Results}
\subsection{Experimental Settings}

\textbf{Datasets.} Three publicly available video re-ID datasets are used for evaluation: PRID-2011 \cite{prid2011}, iLIDS-VID \cite{eccv14video} and MARS \cite{eccv16mars} dataset.
The PRID-2011 dataset is collected from two disjoint surveillance cameras with significant color inconsistency. It contains 385 person video tracks in camera A and 749 person tracks in camera B. Among all persons, 200 persons are recorded in both camera views. Following \cite{cvpr16top,ijcai16video,iccv15des,eccv16mars}, 178 person video pairs with no less than 27 frames are employed for evaluation.
iLIDS-VID dataset is captured by two non-overlapping cameras located in an airport arrival hall, 300 person videos tracks are sampled in each camera, each person track contains 23 to 192 frames.
MARS dataset is a large scale dataset, it contains 1,261 different persons whom are captured by at least 2 cameras, totally 20,715 image sequences achieved by DPM detector and GMCCP tracker automatically.

\textbf{Feature Extraction.} The hand-craft feature LOMO \cite{cvpr15lomo} is selected as the frame feature on all three datasets. LOMO extracts the feature representation with the Local Maximal Occurrence rule. All the image frames are normalized to $128 \times 64$. The original 26960-dim features for each frame are then reduced to a 600-dim feature vector by a PCA method for efficiency considerations on all three datasets. Meanwhile, we conduct a max-pooling for every 10 frames to get more robust video feature representations.

\textbf{Settings.} All the experiments are conducted following the evaluation protocol in existing works \cite{ijcai16video,cvpr16top}. PRID-2011 and iLIDS-VID datasets are randomly split by half, one for training and the other for testing. In testing procedure, the regularized minimum set distance \cite{yang2013face} of two persons is adopted. Standard cumulated matching characteristics (CMC) curve is adopted as our evaluation metric. The procedure are repeated for 10 trials to achieve statistically reliable results, the training/testing splits are originated from \cite{cvpr16top}. Since MARS dataset contains 6 cameras with imbalanced tracklets in different cameras, we initialize the tracklets in camera 1 as the base graph, the same number of tracklets from other five cameras are randomly selected to construct a graph for matching. The evaluation protocol on MARS dataset is the same as \cite{eccv16mars}, CMC curve and mAP (mean average precision) value are both reported.

\textbf{Implementation.} Both the graph matching and metric learning optimization problems can be solved separately using existing methods. We adopt Hungarian algorithm to solve the graph matching problem for efficiency considerations, and metric learning method (MLAPG) in \cite{iccv15liao} as the baseline methods. Some advanced graph matching and metric learning methods may be adopted as alternatives to produce even better results as shown in Section \ref{sec:exp_ml}. We report the results at $10th$ iteration, with $\lambda =0.5$ for all three datasets if without specification.
\begin{figure}[t]
  \centering
  \begin{minipage}[t]{0.25\textwidth}
        \centering
        \includegraphics[width=4.5cm,height=4.5cm]{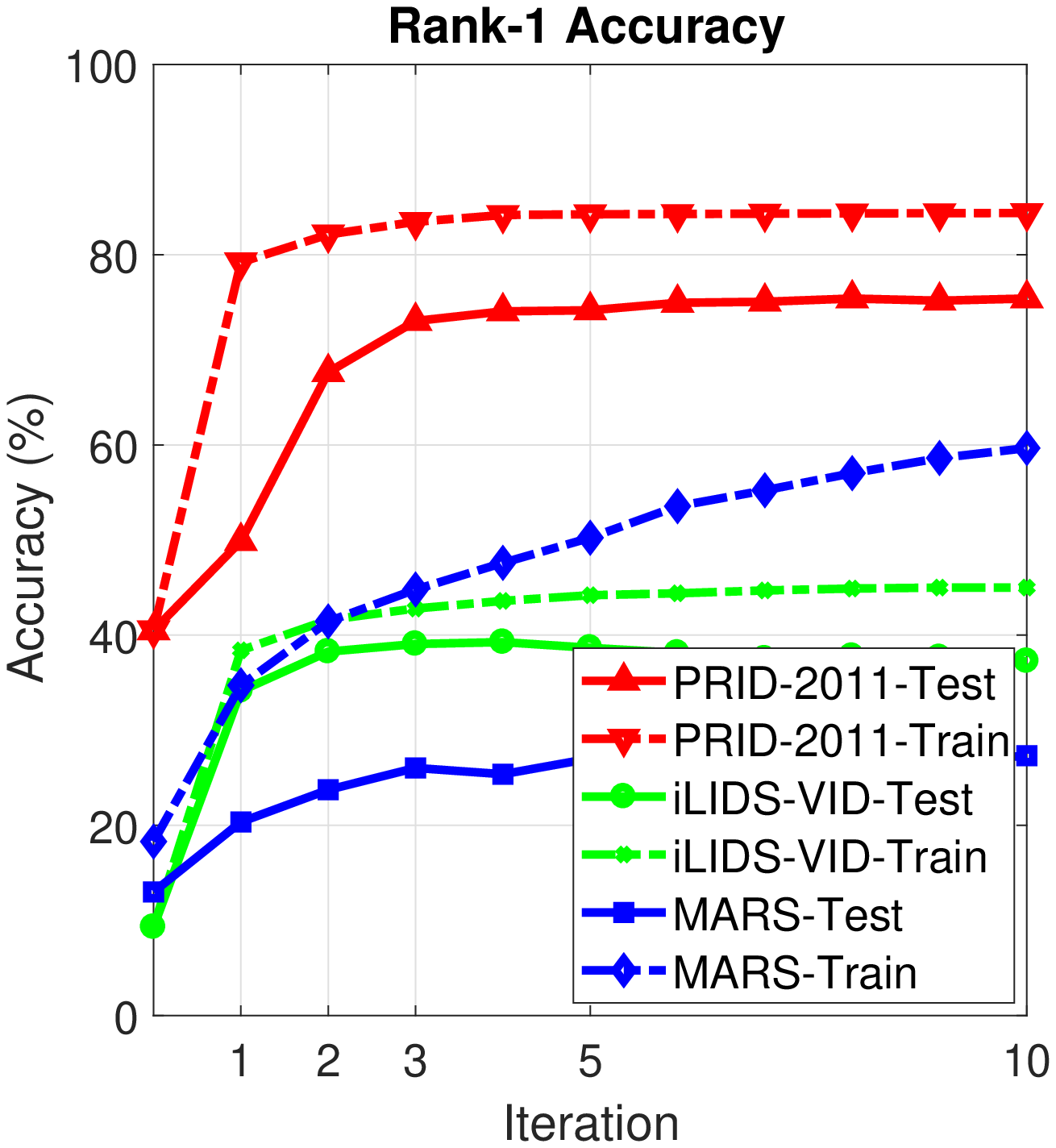}
        \centerline{\small{(a) } }\medskip
    \end{minipage}%
    \begin{minipage}[t]{0.25\textwidth}
        \centering
        \includegraphics[width=4.5cm,height=4.5cm]{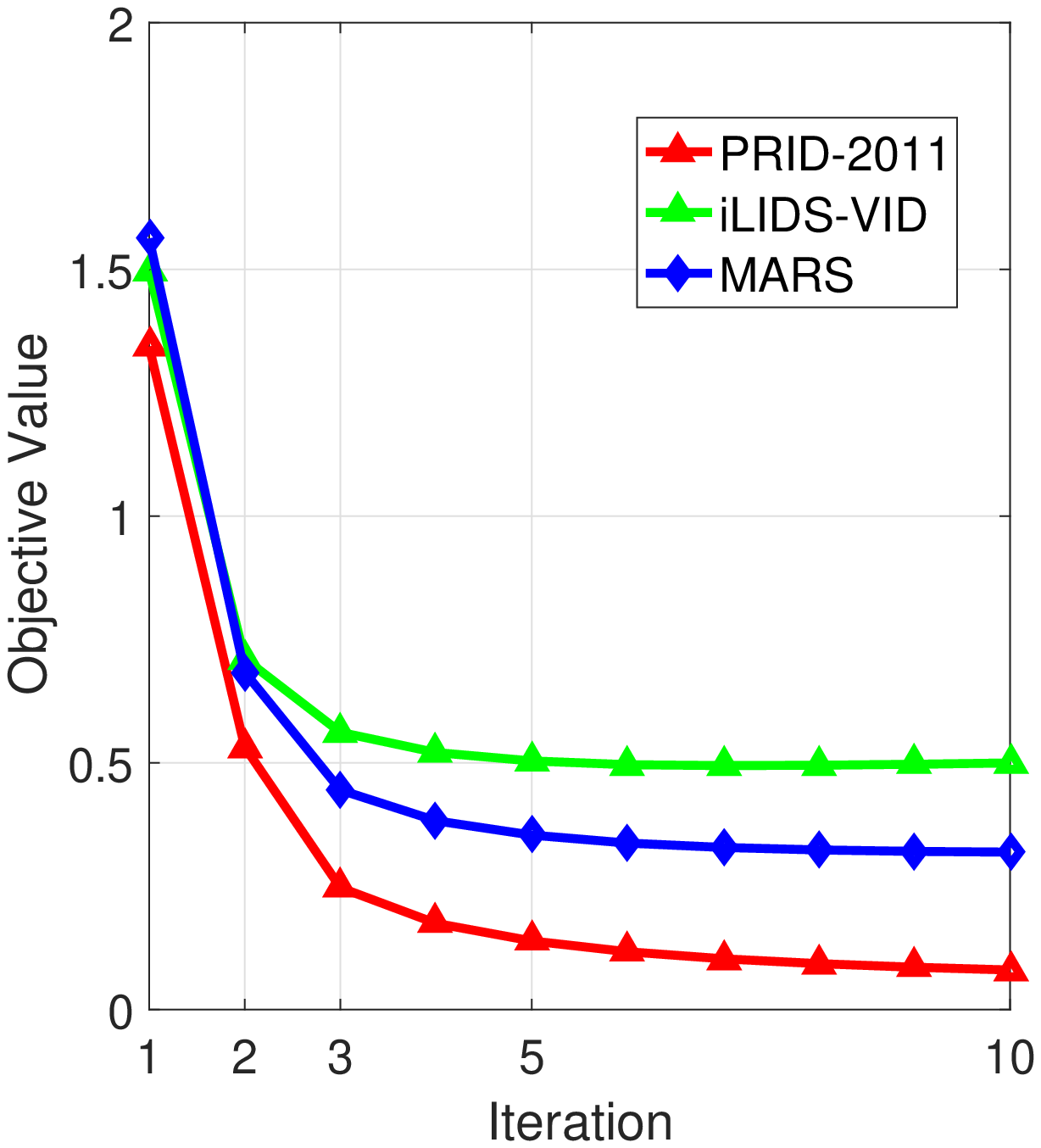}
        \centerline{\small{(b) }}\medskip
    \end{minipage}%
\caption{\small{(a) Rank-1 accuracy of training and testing at each iteration on three datasets. (b) Overall objective values at each iteration on three datasets. For better view, the objective values are normalized.}}\label{fig:iter}
\vspace{-0.1cm}
\end{figure}
\subsection{Self Evaluation}\label{sec:self}

\noindent{\textbf{Evaluation of iterative updating.}}
To demonstrate the effectiveness of the iterative updating strategy, the rank-1 matching rates of training and testing at each iteration on three datasets are reported in Fig. \ref{fig:iter}. Specifically, the rank-1 accuracy for testing is achieved with the learnt metric at each iteration, which could directly reflect the improvements for re-ID task. Meanwhile, the overall objective values on three datasets are reported.

Fig. \ref{fig:iter}(a) shows that the performance is improved with iterative updating procedure. We could achieve 81.57\% accuracy for PRID-2011, 49.33\% for iLIDS-VID and 59.64\% for MARS dataset. Compare with iteration 1, the improvement at each iteration is significant. After about 5 iterations, the testing performance fluctuates mildly. This fluctuation may be caused by the data difference of the training data and testing data. It should be pointed out that there is a huge gap on the MARS dataset, this is caused by the abundant distractors during the testing procedure, while there is no distractors for training \cite{eccv16mars}.
Experimental results on the three datasets show that the proposed iterative updating algorithm improves the performance remarkably. Although without theoretical proof, it is shown in Fig. \ref{fig:iter}(b) that DGM converges to steady and satisfactory performance.

\textbf{Evaluation of label re-weighting.}
We also compare the performance without label re-weighting strategy. The intermediate labels output by graph matching are simply transformed to $1$ for matched and $-1$ for unmatched pairs. The rank-1 matching rates on three datasets are shown Table \ref{tab:reweigh}. Consistent improvements on three datasets illustrate that the proposed label-re-weighting scheme could improve the re-ID model learning.  
\begin{table}[t]\small
\centering
  \begin{tabular*}{\columnwidth}{l|ccc}
  \hline
   Datasets & PRID-2011  &iLIDS-VID  & MARS    \\\hline
  w/o re-weighting  & 72.6     &35.6  & 22.8 \\
  w re-weighting    & 73.1  &37.1   & 24.6   \\ \hline
 \end{tabular*}
 \caption{\label{tab:reweigh}\small{Rank-1 matching rates with (/without) label re-weighting on three datasets.}}
\end{table}

\textbf{Evaluation of label estimation.}
To illustrate the label estimation performance, we adopt the general precision, recall and F-score as the evaluation criteria. The results on three datasets are shown in Table \ref{tab:label}. Since graph matching usually constrains full matching, the precision score is quite close to the recall on the PRID-2011 and iLIDS-VID datasets. Note that the precision score is slightly higher than recall is due to the proposed positive re-weighting strategy.
\begin{table}[t]\small
\centering
\setlength{\tabcolsep}{11pt}
  \begin{tabular*}{\columnwidth}{l|ccc}
  \hline
  Dataset  & Precision   & Recall & F-score \\\hline
  PRID2011  & 82.14     &81.57  & 81.85 \\
  iLIDS-VID & 49.33     &48.64  & 48.98 \\
  MARS      & 59.64     &42.40  & 49.57 \\ \hline
 \end{tabular*}
 \caption{\label{tab:label}\small{Label estimation performance (\%) on three datasets.}}
\end{table}
\begin{figure*}[t]
  \centering
  \includegraphics[width=17cm, height=3.8cm]{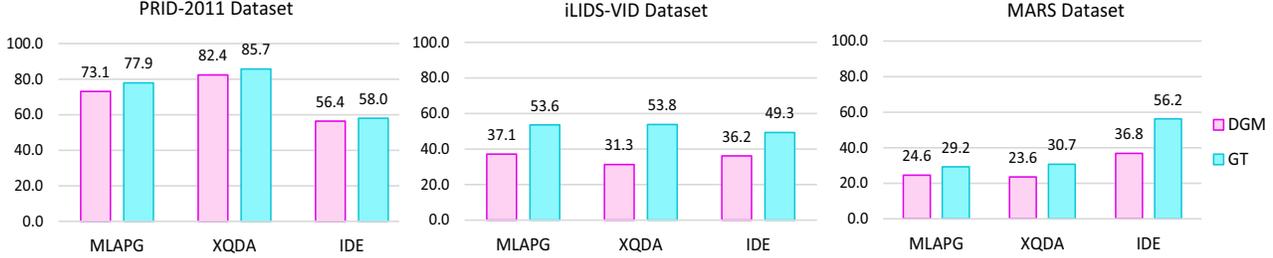}
\caption{\small{Estimated labels for other supervised learning methods. ``DGM" represents the re-identification performance with our estimated labels. ``GT" provides upper bounds with fully supervised learning. Rank-1 matching rates (\%) are reported for three datasets.}}\label{fig:super}
\vspace{-0.1cm}
\end{figure*}

\textbf{Running time.}
The running times on three datasets with the settings described in Section 5.1 are evaluated. It is implemented with Matlab and executed on a desktop PC with i7-4790K @4.0 GHz CPU and 16GB RAM. The training and testing time are reported by the average running time in 10 trials. For training, since we adopt an efficient graph matching algorithm and accelerated metric learning \cite{iccv15liao}, the training time is acceptable. The training time for the PRID2011 dataset is about 13s, about 15s for iLIDS-VID dataset, about 2.5 hours for the MARS dataset due to the large amount of tracklets. For testing, the running time is fast for our method, since standard 1-vs-N matching scheme is employed. The testing times are less than 0.001s on PRID2011 and iLIDS-VID datasets for each query process, and around 0.01s on MARS with 636 gallery persons.
\subsection{Estimated Labels for Supervised Learning}\label{sec:exp_ml}
This subsection evaluates the effectiveness of the output estimated labels for other supervised learning methods. Compared with the re-identification performances with groundtruth labels (GT), they provide upper bounds as references to illustrate the effectiveness of DGM.
Specifically, two metric learning methods MLAPG \cite{iccv15liao} and XQDA \cite{cvpr15lomo}, and an ID-discriminative Embedding (IDE) deep model \cite{eccv16mars} are selected for evaluation as shown in Fig. \ref{fig:super}.

Configured with MLAPG and XQDA, the performances outperform the baseline $l_2$-norm on all three datasets, usually by a large margin. The results show that the estimated labels also match well with other supervised methods. Compared with the upper bounds provided by supervised metric learning methods with groundtruth labels, the results on PRID-2011 and MARS datasets are quite close to the upper bounds. Although the results on iLIDS-VID dataset are not that competitive, the main reason can be attributed to its complex environment with many background clutters, such as luggage, passengers and so on, which cannot be effectively solved by a global descriptor (LOMO) \cite{cvpr15lomo}.

Another experiment with IDE deep model on the three datasets shows the expendability of the proposed method to deep learning methods. Specifically, about 441k out of 518k image frames are labelled for 625 identities on the large scale MARS dataset, while others are left with Eq. \ref{eq:rew}. The labelled images are then resized to $227 \times 227$ pixels as done in \cite{eccv16mars}, square regions $224 \times 224$ are randomly cropped from the resized images. Three fully convolutional layers with 1,024, 1,024 and $N$ blobs are defined by using AlexNet \cite{nips12alex}, where $N$ denotes the labelled identities on three datasets. The FC-7 layer features (1,024-dim) are extracted from testing frames, maxpooling strategy is adopted for each sequence \cite{eccv16mars}. Our IDE model is implemented with MxNet. Fig. \ref{fig:super} shows that the performance is improved with a huge gap to hand-craft features with deep learning technique on the large scale MARS dataset. Comparably, it does not perform well on two small scale datasets (PRID-2011 and iLIDS-VID dataset) compared to hand-craft features due to the limited training data. Meanwhile, the gap between the estimated labels to fully supervised deep learning methods is consistent to that of metric learning methods. Note that since one person may appear in more than one cameras on the MARS dataset, the rank-1 matching rates may be even higher than label estimation accuracy.
\subsection{Comparison with Unsupervised re-ID}

\begin{table*}[t]\small
\centering
\setlength{\tabcolsep}{7.1pt}
 \begin{tabular}{l|cccc|cccc|cccc|c}
  \hline
  Datasets &\multicolumn{4}{c|}{PRID-2011}  & \multicolumn{4}{c|}{iLIDS-VID}                          & \multicolumn{5}{c}{MARS}\\ \hline
  Rank at $r$  & $1$  &$5$   & $10$    & $20$       & $1$  &$5$   & $10$    & $20$       & $1$  &$5$   & $10$    & $20$ &mAP \\\hline
  L2            & 40.6  &66.7    &79.4       &92.3            & 9.2  & 20.0 &27.9  &46.9                 & 14.9  & 27.4  & 33.7  &40.8 & 5.5\\ \hline
  FV3D \cite{iccv15des}&38.7 &71.0 &80.6 &90.3                                 & 25.3 &54.0 &68.3 &\color{red}{\textbf{87.3}}                  & -  & -  & -  &- & -\\
  STFV3D$^{*}$ \cite{iccv15des} & 27.0 &54.0    &66.3  &80.9      & 19.1 & 38.8  &51.7  &70.7            & -  & -  & -  &- &- \\
  Salience \cite{cvpr13saliency}&25.8 &43.6&52.6    &62.0     &10.2 &24.8   &35.5 & 52.9                 & - &- &- &-& -\\
  DVDL \cite{iccv15}    & 40.6  &69.7  &77.8 &  85.6          & 25.9  & 48.2   & 57.3   & 68.9          & -  & -  & -  &- & -\\
  GRDL \cite{eccv16un} & 41.6  &76.4    &84.6   &89.9         &25.7 &49.9   &63.2 & 77.6                &19.3   & 33.2 & 41.6 & 46.5 & 9.56\\
  UnKISS \cite{avss16reid}  & 58.1  &81.9  &89.6  &96.0       & 35.9  &\color{red}{\textbf{63.3}} &\color{red}{\textbf{74.9}} & \color{blue}{\textbf{83.4}}                   & 22.3 &37.4 &47.2 &53.6& 10.6\\ \hline \hline
    DGM + MLAPG \cite{iccv15liao}        & \color{blue}{\textbf{73.1}}  &\color{blue}{\textbf{92.5}} & \color{blue}{\textbf{96.7 }}  & \color{blue}{\textbf{ 99.0}}                  & \color{red}{\textbf{37.1}}  &61.3 & 72.2   &82.0             & \color{blue}{\textbf{24.6 }}& \color{blue}{\textbf{42.6}} &\color{blue}{\textbf{50.4}} &\color{blue}{\textbf{57.2}} & \color{blue}{\textbf{11.8}}\\
  DGM + XQDA \cite{cvpr15lomo}         & \color{red}{\textbf{82.4}}  &\color{red}{\textbf{95.4}} & \color{red}{\textbf{98.3 }}  &\color{red}{\textbf{ 99.8}}                  & 31.3  &55.3 & 70.7  &\color{blue}{\textbf{83.4}}           &23.6 & 38.2 &47.9 &54.7 &11.2\\
 DGM + IDE \cite{eccv16mars}        & 56.4  &81.3 &88.0   &96.4                 & \color{blue}{\textbf{36.2}}  &\color{blue}{\textbf{62.8}} & \color{blue}{\textbf{73.6}}   &82.7             & \color{red}{\textbf{36.8 }}& \color{red}{\textbf{54.0}} &\color{red}{\textbf{61.6}} &\color{red}{\textbf{68.5}} & \color{red}{\textbf{21.3}}\\
  \hline
 \end{tabular}
 \caption{\label{tab:compare}\small{Comparison with state-of-the-art unsupervised methods including image and video based methods on three datasets. {\color{red}Red} indicates the best performance while {\color{blue}Blue} for second best.}}
\end{table*}

This section compares the performances to existing unsupervised re-ID methods. Specifically, two image-based re-ID methods, Salience \cite{cvpr13saliency} results originated from \cite{eccv14video}, and GRDL \cite{eccv16un} is implemented by averaging multiple frame features in a video sequence to a single feature vector. Four state-of-the-art unsupervised video re-ID methods are included, including DVDL \cite{iccv15}, FV3D \cite{iccv15des}, STFV3D \cite{iccv15des} and UnKISS \cite{avss16reid}. Meanwhile, our unsupervised estimated labels are configured with three supervised baselines MLAPG \cite{iccv15liao}, XQDA \cite{cvpr15lomo} and IDE \cite{eccv16mars} to learn the re-identification models as shown in Table \ref{tab:compare}.

It is shown in Table \ref{tab:compare} that the proposed method outperforms other unsupervised re-ID methods on PRID-2011 and MARS dataset often by a large margin. Meanwhile, a comparable performance with other state-of-the-art performances is obtained on iLIDS-VID dataset even with a poor baseline input. In most cases, our re-ID performance could achieve the best performances on all three datasets with the learnt metric directly. We assume that the proposed method may yield better results by adopting better baseline descriptors, other advanced supervised learning methods would also boost the performance further. The advantages can be attributed to two folds: (1) unsupervised estimating cross cameras labels provides a good solution for unsupervised re-ID, since it is quite hard to learn invariant feature representations without cross-camera label information; (2) dynamic graph matching is a good solution to select matched video pairs with the intra-graph relationship to address the cross camera variations.

%

\subsection{Robustness in the Wild}\label{sec:exp_o}
This subsection mainly discusses whether the proposed method still works under practical conditions.

\textbf{Distractors.} In real applications, some persons may not appear in both cameras. To simulate this situation for training, we use the additional 158 person sequences in camera A and 549 persons in camera B of PRID-2011 dataset to conduct the experiments. $d\% * N$ distractor persons are randomly selected from these additional person sequences for each camera. They are added to the training set as distractors. $N$ is the size of training set. We use these distractors to model the practical application, in which many persons cannot find their correspondences in another camera.

\textbf{Trajectory segments.} One person may have multiple sequences in each camera due to tracking errors or reappear in the camera views. Therefore, multiple sequences of the same person may be unavoidable to be false treated as different persons. To test the performance, $p\% *N$ person sequences are randomly selected to be divided into two halves in each camera on PRID-2011 dataset. In this manner, about $p\%$ persons would be false matched since the $p\%$ are both randomly selected for two cameras.

Table \ref{tab:exp_o} shows that the performance without one-to-one matching assumption is still stable, with only a little degradation in both situations, this is because: (1) Without one-to-one assumption, it will increase the number of negative matching pairs, but due to the abundant negatives pairs in re-ID task, the influence is not that much. (2) The label re-weighting strategy would reduce the effects of low-confidence matched positive pairs.
\begin{table}[t]\small
\centering
\setlength{\tabcolsep}{12pt}
  \begin{tabular*}{\columnwidth}{l|cccc}
  \hline
  Rank at $r$ & 1  &5  & 10    & 20 \\\hline
  Baseline  & 73.1  &92.5   & 96.7   &99.0\\ \hline
  $d$(\%)    & \multicolumn{4}{c}{Exp 1. Distractors.} \\ \hline
  20       & 72.1  &91.9 & 95.8   & 98.4\\
  50       & 70.3 &90.9 & 95.2  & 98.2\\ \hline
  $p$(\%)    &\multicolumn{4}{c}{Exp 2. Trajectory Segments.} \\ \hline
  20       & 72.3  &92.1 & 95.9   & 98.6\\
  50       & 71.1  &91.6 & 95.4   & 98.3\\ \hline
 \end{tabular*}
 \caption{\label{tab:exp_o}\small{Matching rates (\%) on the PRID-2011 dataset achieved by the learnt metric without one-to-one matching assumption.}}
\end{table}

\section{Conclusion}
This paper proposes a dynamic graph matching method to estimate labels for unsupervised video re-ID. The graph is dynamically updated by learning a discriminative metric. Benefit from the two layer cost designed for graph matching, a discriminative metric and an accurate label graph are updated iteratively. The estimated labels match well with other advanced supervised learning methods, and superior performances are obtained in extensive experiments. The dynamic graph matching framework provides a good solution for unsupervised re-ID.

\textbf{Acknowledgement} This work is partially supported by Hong Kong RGC General Research Fund HKBU (12202514), NSFC (61562048). Thanks Guangcan Mai for the IDE implementation.

{\small
\bibliographystyle{ieee}
\bibliography{ref}
}

\end{document}